# A General Theory for Training Learning Machine

Hong Zhao (zhaoh@xmu.edu.cn)

Department of Physics, Xiamen University

Though the deep learning is pushing the machine learning to a new stage, basic theories of machine learning are still limited. The principle of learning, the role of the *a prior* knowledge, the role of neuron bias, and the basis for choosing neural transfer function and cost function, etc., are still far from clear. In this paper, we present a general theoretical framework for machine learning. We classify the prior knowledge into common and problem-dependent parts, and consider that the aim of learning is to maximally incorporate them. The principle we suggested for maximizing the former is the design risk minimization principle, while the neural transfer function, the cost function, as well as pretreatment of samples, are endowed with the role for maximizing the latter. The role of the neuron bias is explained from a different angle. We develop a Monte Carlo algorithm to establish the input-output responses, and we control the input-output sensitivity of a learning machine by controlling that of individual neurons. Applications of function approaching and smoothing, pattern recognition and classification, are provided to illustrate how to train general learning machines based on our theory and algorithm. Our method may in addition induce new applications, such as the transductive inference.

## 1.Introduction

In essence, a learning machine is a high-dimensional map from input $x \in R^M$ to output $y \in R^L$, $y = \Phi(\Lambda, x)$, where $\Lambda$ represents the parameter set. The goal of the design is to find a parameter set $\Lambda$ in the condition of a given set $(x^\mu, y^\mu, \mu = 1,..., P)$ of $P$ samples, which guarantees the map not only correctly response to the sample set, but also correctly response to the real set represented by the samples. Generalizing the knowledge learned from the limited samples to the real set is the soul of a leaning machine. This ability lies on the *a prior* knowledge being incorporated into the learning machine. Prior knowledge is the information about the learning task which is available in addition to the training samples. A theory for designing a learning machine is therefore usually composed of two parts, i.e., an algorithm for training the learning machine response to samples and strategies for gaining the generalization ability.

There are two basic algorithms for establishing the desired input-output (I-O) responses in designing multilayer learning machines, i.e., the back-propagation (BP) method [1-3] and the support vector machine (SVM) method [4-5]. Both methods have achieved great success in various applications ranging from the traditional domain of function approach, pattern recognition and time-series prediction to an increasingly wide variety of biological, chemical, social science applications [1-10], as well as quantum computation [11-13]. They are also elementary algorithms in designing more complex learning machines such as deep-learning neural networks [14,15].

The BP method employs a set of deterministic equations to calculate the system parameters in an iterative manner to train a neural network response to samples. Such a response is approached by minimizing the empirical risk. To gain the generalization ability, the BP method applies the 'Occam razor principle': the entities should not be multiplied beyond the necessity. Based on this principle, the best machine should have as smaller as possible size. Large size network than necessary is supposed to prone to over-fit the data. However, it is already clear that smaller machines may not always give the best performance [4]. In addition, the BP algorithm is experience dependent, particularly for choosing a proper learning rate.

The SVM method gives up the Occam razor, and follows the structural risk minimization principle of statistical learning theory [4] to gain the generalization ability. Based on the principle, the machine with the smallest Vapnik-Chervonenkis dimension instead of the smallest size is supposed to be the best. For this goal, the SVM method maps the input vectors of samples into the high-dimensional feature space by so-called support vectors of selected samples, and then the feature vectors are separated with the maximum margin hyperplane calculated following the linear optimization algorithm. However, fundamentally, whether the maximum margin hyperplane can generally lead to the best machine is unknown[4,8]. Technically, how to calculate the Vapnik-Chervonenkis dimension and how to choose the kernel function are open problems [8]. In addition, support vectors are input vectors of selected samples, which may result in serious restriction when only a small training set is available. Another drawback is that the SVM is originally developed for binary decision problems. Though there are efforts to extend the method to multi-classification problems, it seems that the performance of such a machine is still poorer than the bi-classification SVMs [16].

Besides the basic principle, a theoretical understanding of roles of the neural transfer function or kernel function, the neuron bias and the cost function seems still absent. In BP based algorithms, the sigmod and logistic functions are usually applied, while in recent years the rectified linear unit function becomes more and more popular. In SVM method, more neural transfer functions are applied in terms of the kernel function, include the Gauss kernel, polynomial kernel, and various kernels constructed for specific applications. The cost function used in the BP algorithm is usually the means square of output errors, while in SVM the so-called hard and soft- margins are both candidates. The choice of these functions seems mainly based on intuition. A systematic theoretical understanding of why a specific choice is better seems currently out of reach.

In this paper, we present a new theory to design learning machines. Examples in the present paper are restricted to architectures with input, hidden, and output layers, but the extension to more complicated learning machines is possible. We keep the architecture as used by the BP method but design the machine based on the statistical learning theory [4]. To train the learning machine responses to samples, we develop a Monte Carlo (MC) algorithm. In fact, Hagan, Demuth and Beale have pointed out that randomly searches for suitable weights may be a possible way in developing their BP algorithm [2]. It was however abandoned because they did not believe it is practicable because of the computational difficulty. We recur this idea since firstly it is flexible enough in finding solutions of optimization problems with complex restrictions. Secondly and essentially, the speed of nowadays' computers has been greatly improved. Another equally important factor is that our algorithm evolves only a small part of the learning machine. At each round of adaptation, we adapt only one parameter if it does not make the performance bad, instead of developing the entire system. Due to this reason, the training time is acceptable for usual applications. Indeed, the MC based algorithms have been extensively applied in machine learning. The procedure of stochastic gradient descent used by most practitioners of deep learning [14], and the algorithm of celebrated AlphaGo game [15] are typical examples benefitted from stochastic algorithms. The mechanism why these stochastic heuristic processes often find solutions has been [17] revealed in terms of a nonequilibrium statistical physics framework. In our approach, the high flexibility of MC algorithm enables one to adopt various neuron transfer functions and cost functions, and to design learning machines with either continuous or discrete system parameters. Since the weight vectors are chosen from the general vector space, we call the learning machine designed by our method the general vector machine (GVM).

To maximize the generalization ability, we classify the prior knowledge into the common and problem-dependent classes, and suggest corresponding strategies to integrate them into the learning machine. The common prior knowledge is further classified into two types. *Objects with small difference in features should belong to the same class* is considered as the common prior knowledge for pattern recognition and classification, and *a natural curve should have sufficient smoothness* is considered to be that for function approach and smoothing, since based on which human beings generalize experiences. To incorporate this type of prior knowledge, a learning machine should be insensitive to small input changes, i.e., should have small I-O sensitivity. This kind of prior knowledge is for objects. Meanwhile, there is a general requirement for a learning algorithm: Giving the same training set, learning machines designed by a good learning algorithm should have as small as possible output uncertainty. We call the uncertainty the design risk. This kind of *a prior* knowledge is not explicitly incorporated in conventional learning methods. We present the design risk minimization (DRM) strategy -- Learning machines with smaller design risk have better performance, as a basic principle. The reason is that, as will be shown by our analysis and examples, apart from being a basic requirement, control the design risk can monotonously control the I-O sensitivity (but not vice versa), and therefore can maximally incorporate both types of common prior knowledge.

We refer the particular background knowledge for particular problems, such as the geometric symmetry of patterns, the physical interpretation of input vectors of samples, the goal function creates the data set, etc., to be the problem-dependent prior knowledge. To maximally incorporate this kind of prior knowledge, one needs to apply individualized strategies, including proper pretreatment of sample input vectors, using proper neuron transfer functions and cost functions, etc..

We introduce a set of parameters to be control parameters, including parameters specifying range of weights, of neural transfer coefficients, of neuron biases, as well as the number of hidden-layer neurons and the width of the separation margin. Instead of finding the best learning machine as conventional methods do, we search for the best control-parameter set. Generally, the best control-parameter set has the smallest design risk. However, for pattern recognition, extremely minimizing the design risk may induce divergence from the pattern's natural symmetry and decreases the recognition accuracy. Therefore, for pattern recognition beside the design risk, we employ also the average recognition accuracy on the test set to identify the best control-parameter set. For calculating the design risk and the average recognition accuracy, we need to design a sufficient amount of GVMs at a control parameter set for a given training set. The MC algorithm is particular suitable for such a task. By performing the training with different random initial realizations in the parameter ranges specified by the control-parameter set, we can obtain an arbitrary amount of statistically identical GVMs.

Every GVM designed at the best control parameter set can be equivalently applied as the performing learning machine since there are statistically identical. Indeed, the strategy that picking up a learning machine having the highest correct rate on the test set is unreasonable and misleading. In other words, designing a large number of statistically identical learning machines indeed necessary for judging any learning method.

GVMs obtained at a control parameter set can be applied to construct a more effective performing learning machine -- a joint GVM (J-GVM). The J-GVM can dramatically decrease the design risk as well as the I-O insensitivity. It may achieve a fair balance between the goal of maximally extracting the feature of input vectors and the goal of minimizing the risks, and thus may significantly improve the generalization ability. The idea of J-GVM is similar to the ensemble method [18]. The difference is that, besides being designed under the supervision of the DRM strategy, the GVMs used to construct the J-GVM are trained by the same training set.

The rest of the paper is managed as follows. In the next section, we introduce the architecture of the GVM. We shall emphasize the difference from that of the SVM. In section 3, we present the MC algorithm for training a GVM in response to samples. Section 4 is provided to introduce the idea of controlling the I-O sensitivity of a learning machine by controlling the second derivatives of single neurons. The main control parameters are introduced in this section. Section 5 introduces in detail the DRM principle and explains why it can maximize the common prior knowledge. Section 6 introduces several strategies for maximally incorporating the problem-dependent prior knowledge. This section presents the theory of choosing the neural transfer function and the cost function. Section 7 shows how to construct the J-GVM. The next three sections are applications, including the function approach and smoothing, pattern recognition and classification, respectively. The concluding section is to summarize the main ideas and results. A particular application, 'washing' out the bad samples using our method, is demonstrated in the end of this section.

## 2. The model

We study the three-layer learning machine composed by input, hidden and output layers. The numbers of neurons in the input, hidden, and output layers are N, M, and L correspondingly. The dynamics is given by the following formula,

Hidden layer:

$$\bar{y}_i = f_i(\beta_i \bar{h}_i), \bar{h}_i = \sum_{j=1}^{M} \overline{w}_{ij} x_j - b_i, i = 1, ..., N \qquad (1)$$

Output layer:

$$y_l = h_l, h_l = \sum_{i=1}^{N} w_{li} \bar{y}_i, l = 1, ..., L \qquad (2)$$

where $\bar{y}_i, f_i, \bar{h}_i, \beta_i, b_i$ respectively represent the output, neuron transfer function, local field, transfer function coefficient, and bias of the *i*th neuron in the hidden layer. Here $x_j$ is the *j*th component of an input vector, $\bar{w}_{ij}$ is the weight connecting the input $x_j$ and the *i*th neuron in the hidden layer. Similarly, $y_l, h_l$ are the output and the local field of the *l*th neuron in the output layer, $w_{li}$ is the weight connecting $\bar{y}_i$ and the *l*th neuron in the output layer, M is the dimension of the input vector, N is the number of neurons in the hidden layer, L is the dimension of output vectors. We apply the linear transfer function to the output layer for simplifying the analysis. For function approach, it is applicable directly. For pattern recognition and classification, one can further apply nonlinear transfer functions to assign labels to output vectors after finishing the training.

Let us denote the *i*th row of the weight matrix $\bar{w}$ by $\bar{w}^i \in R^M$, and call it as the *i*th weight vector of the matrix. The essential difference from the SVM is that we use weight vectors from the general vector space to replace support vectors of selected samples. The architecture of GVMs is the same as that of neural networks designed by the BP method. Nevertheless, in a GVM, not only the weights connecting different neurons but also the neural transfer function coefficients are adjustable parameters. The parameters are found by the MC algorithm presented in the next section.

## 3. Monte Carlo algorithm

The MC algorithm is for establishing the correct response on the training set. A three-layer network maps input vectors to output vectors by two steps of transformation. That is: the hidden layer maps the M-dimension input vectors of samples into N-dimension vectors in the 'feature space', and then the output layer maps them into L-dimensional output vectors. The two layers resemble two coupled mirrors. Simultaneously changing two mirrors, or fixed on one and changing another, could both establish the desired input-output correspondence. The SVM algorithm applies the former strategy. It sets the hidden-layer by support vectors of selected samples, and calculates the output-layer by the linear optimization theory. Introducing the support vectors is the soul of the SVM method. It decreases dramatically freedom for choosing the hidden-layer parameters, and reduces the solution to be a linear optimization problem in the feature space. This treatment, on the other hand, imposes restriction on feature extraction since the support vectors can only be chosen from input vectors of samples.

Our idea is different. We give up the support vectors and apply directly the weight vectors to collect the feature of input vectors. To perform the training, we randomly initialize the parameters in the output layer and fixed them afterwards, and then adjust the 'hidden-layer mirror', which is established by the weight vectors, neuron transfer function coefficients and neuron biases, to find the solution. Because there is a huge amount of parameters in the hidden layer, the possibility to find solutions with the fixed output layer is still much high. One can in principle adjust simultaneously parameters in both layers to find the solution. However, once the 'output-layer mirror' changes, the 'hidden-layer mirror' should adjust accordingly to match the change, which may significantly increase the training time.

For sake of simplicity, we set random binary values to the weights in the output layer, i.e., $w_{li} = \pm 1, l = 1,...,L, i = 1,...,N$. The parameters in the hidden layer are initialized with random numbers within their available ranges (will be defined in the next section).

To supervise the training, a cost function, $F = F(\,y^{\mu}, t^{\mu}\,), \mu = 1, ..., P$, is constructed by training samples, where $t^{\mu}$ represents the actual output of the learning machine under the input $x^{\mu}$. In sections 5 and 6, we shall show how to construct $F$.

To start the training, we first calculate the local fields of neurons, $\bar{h}_i^{\mu}, h_l^{\mu}, \mu = 1, ..., P; i = 1, ..., N; l = 1, ..., L$, as well as the function $F$, for a given training set. Note that $y^{\mu} = h^{\mu}$ by applying the learning transfer function. We then repeatedly apply the following procedure to adjust hidden-layer parameters: Randomly adapt one hidden-layer parameter to a new value in its available range, and calculate the changes in $F$; If it does not become worse, accept the adaptation and renew the local fields, the outputs of neurons and the cost function, otherwise give up the adaptation. The hidden layer is renewed by the follow rules:

$$\text{If } \ \bar{w}_{ij} \rightarrow \bar{w}_{ij} + \varepsilon, \ \text{ then } \ \bar{h}_i^{\mu} \rightarrow \bar{h}_i^{\mu} + \varepsilon x_j^{\mu}, \bar{y}_i^{\mu} = f_i(\beta_i \bar{h}_i^{\mu}) \qquad (3)$$

$$\text{If } \ \beta_i \rightarrow \beta_i + \varepsilon, \ \text{ then } \ \bar{h}_i^{\mu} \rightarrow (\beta_i + \varepsilon)\bar{h}_i^{\mu}, \bar{y}_i^{\mu} = f_i(\beta_i \bar{h}_i^{\mu}) \quad (4)$$

$$\text{If } \ b_i \rightarrow b_i + \varepsilon, \ \text{ then } \ \bar{h}_i^{\mu} \rightarrow \bar{h}_i^{\mu} - \varepsilon, \bar{y}_i^{\mu} = f_i(\beta_i \bar{h}_i^{\mu}) \qquad (5)$$

The output layer is renewed by

$$h_l^{\mu} \rightarrow h_l^{\mu} + w_{li}(\bar{y}_i^{\mu} - \bar{y}_i^{\mu}(old)) \qquad (6)$$

where $\bar{y}_i^{\mu}(old)$ represents the value before adaptation. The renewal operations are performed over $\mu = 1, ..., P; l = 1, ..., L$. For a particular application, one can specify only certain classes of parameters as changeable ones and keeps the others fixed. For designing a learning machine with continuous parameters, one can set $\varepsilon$ to be small random numbers. It is not necessary to limit $\varepsilon$ to be sufficiently small, if only the renewed parameter remains still in the corresponding available range. For designing a learning machine with parameters taking only discrete states, as studied in [19,20], the parameter $\varepsilon$ works for pushing the selected parameter jumping from the present state to neighboring states. The training is stopped until $F \leq F_0$ or after a sufficiently long training time, $t \geq t_0$.

Our algorithm does not need to develop the entire network which needs about $O(NMP + NLP)$ multiply-add operations. Each adaptation induces only $O(P + LP)$ multiply-add operations and the adaptation accepted is optimum for the whole training set in the statistical sense. Examples shown in the application sections indicate that the training time is practically available for various applications.

Applying the MC algorithm, feature extraction is approached by projecting an input vector into weight vectors in the hidden layer. The projections are seen as the 'features'. Though a single weight vector may extract less information than a support vector does, the unlimited amount of weight vectors can offset this drawback. Indeed, none has proved that support vectors are the best ones. Particularly, when the training set is quite small, the support vector method has an obvious limitation. More weight vectors means more feature information. As a consequence, we prefer large learning machine to small ones. The over-training problem induced by the large network can be suppressed by controlling the design risk using strategies introduced in next two sections.

# 4. Controlling the I-O sensitivity

In this section, we show how to control the I-O sensitivity of a GVM by controlling the second derivatives of single neurons. The response of a function $y = f(x)$ to a small input change can be expressed as $\delta y \sim \delta x \frac{\partial f}{\partial x} + \frac{1}{2}(\delta x)^2 \frac{\partial^2 f}{\partial x^2} + \cdots$. Therefore, the moments of derivative determine the I-O sensitivity of the function. In application, the second moment is usually applied to characterize the sensitivity. The second moment of the derivative of the $i$th neuron in the hidden layer to the $j$th and $k$th components of an input vector is $\gamma^i_{jk} \equiv \frac{\partial^2 \bar{y}_i}{\partial x_j \partial x_k}$, which can be evaluated explicitly:

$$\gamma^i_{jk} = \beta_i^2 \bar{w}_{ji} \bar{w}_{ki} f_i^{''}(\beta_i \bar{h}_i) \qquad (7)$$

The I-O sensitivity of a GVM is the linear combination of I-O sensitivities of neurons, i.e., for the $l$th output of a GVM it has $\frac{\partial^2 y_l}{\partial x_j \partial x_k} = \sum_{i=1}^{N} w_{il} \gamma^i_{jk}$. It is therefore determined by totally LM$^2$N terms of second derivatives of $\gamma^i_{.jk}$. It is a hard task for calculation, let alone to find the solution of the minimum. For a monitoring purpose, one may employ the average magnitude of

$$R_S = < \left\| \sum_{i=1}^{N} w_{il} \gamma^i_{jk} \right\| > \qquad (8)$$

to estimate the I-O sensitivity of a GVM, where $<\bullet>$ represents the average over index pair $(l,j,k)$.

Following Eq. (7), the second derivatives of a neuron are determined by the product of $\beta_i^2, \overline{w}_{ji}, \overline{w}_{ki}$ and $f_i''$. Therefore, the magnitude of the second derivatives can be controlled by specifying the available ranges of neural transfer coefficients and weights, i.e.,

$$\beta_i \in [-c_\beta, c_\beta], \quad \overline{w}_{jk} \in [-c_w, c_w], \tag{9}$$

with proper neuron transfer functions having bounded derivatives. Typical neuron transfer functions suitable for this requirement include

$$f_i(z) = e^{-z^2} \tag{10}$$

with $f_i'' \in [-2, 0.9]$, and

$$f_i(z) = \tanh(z) \tag{11}$$

with $f_i'' \in [-0.8, 0.8]$. The former is called as the Gauss and the latter the sigmoid neuron transfer functions hereafter. The ReLU function $f_i(z) = \max(0, z)$ has singular derivative and is not a favorable neural transfer function from our criterion.

For some applications, the polynomial function

$$f_i(z) = z^n \tag{12}$$

is also applied as a neural transfer function. Since $f_i''(z) = n(n-1)z^{n-2}$, its amplitude has a linear dependent on the input amplitude, and it increase sharply with the exponent $n$. We therefore need to introduce additional strategy to suppress the I-O sensitivity when applies it.

The parameter $b_i$ influences the second derivatives by influencing $f_i^{''}$. The latter is determined by also $\beta_i$ and weights, as well as the input vectors. Amplitudes of input vectors themselves are expected not to influence the sensitivity of a learning machine. For function approach, at $z=0$ it has $f_i^{''} = f_i^{''}(0)$ for all of the hidden-layer neurons if $b_i=0$. In this case, if $f_i^{''}(0)=0$ as in the cases of Gauss and sigmoid neuron transfer functions, the I-O sensitivity becomes out of control by the control parameters. While with a larger input coordinate value, the amplitude of the local field should be large following Eq. (1), which thus systematically influences $f_i^{''}$. To avoid this singularity, the parameter $b_i$ is given a role in controlling the I-O sensitivity, by assigning random values to the range

$$b_i \in [-c_b, c_b] \tag{13}$$

With $c_b \sim max\sum \overline{w}_{ij} x_j^\mu$. Here without loss of the generality, we suppose that input vectors are distributed around the origin. As $b_i$ and $\overline{w}_{ij}$ take random values, $f_i^{''}$ takes different values even with the zero vector input. In this way, the influence of input vectors is suppressed significantly, and the intrinsic I-O sensitivity can be controlled by $c_\beta$ and $c_w$ effectively.

For pattern recognition, one more control parameter has to be introduced. It is the width of the separating margin. The GVM of pattern recognition has a $M-N-P$ architecture. The input vector $x^\mu$ of the $\mu$th sample represents the $\mu$th pattern. Suppose it belongs to the $\nu$th class. Then the output vector $y^\mu$ is expected to take the components

$$y_l^\mu = \begin{cases} > 0, l = \nu, \\ \leq 0, otherwise. \end{cases} \tag{14}$$

Such a GVM responses correctly to the training set if only $h_l^\mu s_l^\mu > 0$ , for

$\mu = 1,...,P; l = 1,...,L$, where $s_l^\mu = sign(y_l^\mu)$. With this condition satisfied, the empirical risk vanishes. However, such a GVM is sensitive to fluctuations and lack of generalization ability. To make the GVM insensitive to input changes, the condition should be replaced by $h_l^\mu s_l^\mu - d > 0$. A variation of the μth sample, $x = x^\mu + \delta x$, induces an output, $h = h^\mu + \delta h$. This variation is classified into the class that the μth sample belongs, if $h_l^\mu s_l^\mu - d + \delta h_l s_l^\mu > 0$ for all of the components *l*. To guarantee this condition be satisfied for as large as possible variations $\delta x$, one can fix the separation margin and decrease amplitudes of derivatives, or fix the latters and increase the former. Both ways can decrease the I-O sensitivity. Nevertheless, how to quantitatively define and calculate the I-O sensitivity is a hard task.

## 5. Incorporating the common prior knowledge

Let's recall the common prior knowledge: A learning machine should have small I-O sensitivity and learning machines designed by the same training set should have as small as possible output uncertainty. Obviously, small design risk implies small I-O sensitivity. At the minimum of the design risk, the I-O sensitivity of GVMs should be at a low level to maintain such a low design risk. No learning theory has proven that the minimum I-O sensitivity gives the best learning machine. Indeed, our example in the application section 8 will show that extremely minimizing the I-O sensitivity can induce the overtraining. Therefore, we suggest the DRM to be a principle to incorporate both kind of prior knowledge.

For function approach we show that the DRM strategy can generally lead to the best fitting. The goal of function approach is to find a learning machine $\Phi(\Lambda, x)$ satisfying $\Phi(\Lambda, x) = g(x)$, where $g(x)$ is the goal function. The empirical risk

$$F_e = \sqrt{\frac{1}{P}\sum_{\mu=1}^{P}(t^{\mu} - y^{\mu})^2} \, , \tag{15}$$

where $t^{\mu}$ and $y^{\mu}$ are the target and actual output of the μth sample input, is applied as the cost function. To calculate the design risk, we construct n GVMs satisfying $F_e \leq F_c$ with randomly initialized internal parameters at a given control parameter set. We define $\Pi(x) = \Phi(x, \Lambda)$ to be the response function of a GVM. The squared error

$$E(\Pi) \equiv \sqrt{\frac{1}{n}\sum_{i=1}^{n}(\|\Pi_i(x) - \langle\Pi(x)\rangle\|)^2} \tag{16}$$

defines the design risk, where $\Pi_i(x)$ is the response function of the $i$th GVM, $\|\circ\|$ represents the norm of the function, and $\langle\Pi(x)\rangle$ is the ensemble average of response function. With the increase of training samples, $<\Pi(x)>$ should converge to the goal function. In such a case, $E(\Pi)$ converges to the average fitting error defined as

$$\langle\Theta(\Pi)\rangle \equiv \sqrt{\frac{1}{n}\sum_{i=1}^{n}(\|\Pi_i(x) - g(x)\|)^2} \, , \tag{17}$$

and the minimum of the design risk gives that of the average fitting error. Nevertheless, for a finite training set, the convergence cannot be guaranteed in principle. The function $\langle\Pi(x)\rangle$ is determined by training samples and the neural transfer function, as well as the size of the GVM. Despite this is the case, $\langle\Pi(x)\rangle$ should be the optimum approach to the goal function for a given GVM with a specific neuron transfer function, since it get rid of fluctuations. Fortunately, our examples will show that usually $\langle \Pi(x) \rangle$ converges almost to the goal function even for small training sets.

For pattern recognition, it pursues that arbitrary variations $x = x^{\mu} + \delta x$ of $x^{\mu}$ are classified into the class of the μth sample. For this purpose, we formulate the condition $h_l^{\mu} s_l^{\mu} - d > 0$ with following cost function,

$$F_1 = \frac{1}{PL} \sum_{\mu=1}^{P} \sum_{l=1, h_l^{\mu} s_l^{\mu} < d}^{L} (h_l^{\mu} s_l^{\mu} - d)^2 , \qquad (18)$$

where $s_{\nu}^{\mu} = 1$ and $s_l^{\mu} = -1$ for $l \neq \nu$ if the sample belong to the $\nu$th class. When the cost function is minimized to $F_1 = 0$, the local fields of neurons in the output layer satisfy $h_l^{\mu} s_l^{\mu} > d$ for all training samples. Minimizing the I-O sensitivity with this cost function maximizes the common prior knowledge of *Objects with small difference in features should belong to the same class*. The SVM indeed applies such a strategy to gain the generalization ability.

Instead of the I-O sensitivity, our theory pursues a minimum design risk. To calculate the design risk, we employ the correct rate of a GVM on the test set to be the response function $\Pi$. At a given control parameter set, we construct n GVMs by use of the MC algorithm and obtain a response series $\Pi_i, i = 1, ..., n$. With this series we can calculate the average correct rate $\langle \Pi(x) \rangle$ and the dispersion degree of correct rates. The latter defines just the design risk. We then search for the minimum of design risk in the space of control parameters. Similar to function approach, the design risk can also control the I-O sensitivity, and thus incorporate the common prior knowledge either for recognition objects or for recognition systems.

## 6. Incorporating the problem-dependent prior knowledge

Common knowledge adds no restriction to variations. It generally demands to classify a random variation of as big as possible fluctuations to the class that the sample belongs to. In other words, the strategy of maximizing the common prior knowledge is applicable for the case that real patterns can be considered as random variations of samples. For most applications, real patterns are usually not random variations; they are restricted by their own geometry. Problem-dependent prior knowledge therefore must be taken into consideration.

For pattern recognition, there are various types of problem-dependent prior knowledge, such as the interpretation of input vectors and the geometric symmetry of patterns, etc. The physical interpretation of input vectors may also involve the problem-dependent prior knowledge. When a variation $x = x^\mu + \delta x$ of the $\mu$th sample is input to a GVM, deviations $\bar{h}_i \rightarrow \bar{h}_i^\mu + \delta \bar{h}_i$ and $h_i \rightarrow h_i^\mu + \delta h_i$ of local fields in the hidden layer and in the output layer will be induced in turn. In medical diagnosis, for example, the component of an input vector describes a biochemical indicator, being endowed the meanings that the more low the value the more normal, and the more high the more likely malignant. This fact demands that $\bar{h}_i$ and $h_i$ possesses a monotonic relation to $\delta x$. In other examples as for pattern recognition, an input vector encodes a two-dimensional pattern. A component of such a vector has a standard reference value for a specific pattern. The corresponding component of a new pattern with either higher or lower value both represent the deviation from the reference value. In this situation, the single-peak functions match the feature better.

We suggest adopting proper neural transfer function and cost function that matches the particular prior knowledge better to improve the learning machine performance. For example, the sigmoid transfer function should be preferable than the Gauss transfer function for the example of medical diagnosis, while the situation would be reversed in the case of geometry pattern recognition. The cost function can be chosen similarly. The cost function $F_1$ defined by Eq. (18) matches the first example better. It has $h_\nu^\mu > d$ and $h_l^\mu < d, l \neq \nu$ otherwise. While we can also construct the cost function in another way,

$$F_2 = \frac{1}{PL} \sum_{\mu=1,}^{P} \sum_{l=1}^{L} (h_l^\mu s_l^\mu - d)^2 \tag{19}$$

which compresses $h_l^\mu s_l^\mu$ around $d$. Obviously, this function matches the second class of examples better.

In certain case, $F_2$ may not be approachable due to the huge amount of training samples. In this situation, introducing the following cost function,

$$F_3 = \frac{1}{PL} \sum_{\mu=1}^{P} \begin{cases} \displaystyle\sum_{l=1, h_l^\mu s_l^\mu < d_1}^{L} (h_l^\mu s_l^\mu - d_1)^2 \\ \displaystyle\sum_{l=1, h_l^\mu s_l^\mu > d_2}^{L} (h_l^\mu s_l^\mu - d_1)^2 \end{cases} \tag{20}$$

may be helpful. The minimization of it drives $h_l^\mu s_l^\mu$ into the interval $[d_1, d_2]$. When $d_2 >> d_1$ it approaches $F_1$, while when $d_2 \sim d_1$ it approaches $F_2$.

These cost functions can also be presented in another way. For example, the second cost function can be modified as

$$\overline{F}_2 = \frac{1}{PL} \sum_{\mu=1}^{P} \sum_{l=1, l \neq \nu}^{L} (h_\nu^\mu - h_l^\mu - d)^2 \qquad (21)$$

By minimizing this function, $h_\nu^\mu$ and $h_l^\mu$ for $l \neq \nu$ are separated by a distance $d$ for each sample patterns, but the local fields need not to distribute around the origin.

Proper pretreatment of input vectors is also an effective manner to incorporate the problem-dependent knowledge of transformation symmetry. For handwritten digit recognition, patterns have symmetries under small spatial shifts, rotations, as well as distortions. For man-made objectives, spatial shifts and angle variations are important but distortions may have no corresponding. For these particular problems, generalization based on random variations is extravagant. To incorporate the specific symmetric restriction, one can construct spurious training samples by the shift, rotation, distortion, tangent distance technique, etc. [21], and applies them also in the training. The drawback is that a big amount of additional calculations is arisen for training the machine. Another way is to encode the symmetry property into the input vectors of samples. The so-called Gradient-base feature-extraction algorithm [22], with feature vector encoding eight direction-specific $5 \times 5$ gradient images, is one of the top-performing algorithms for this purpose.

To identify whether the problem-dependent prior knowledge is incorporated, we have to apply also the average correct rate on the test set to be another performance indicator beside the design risk. The correct rate will increase when a proper strategy or treatment is applied.

Furthermore, it can also identify the turning point when extremely maximizing the common prior knowledge leads to the deviation from the natural symmetry of patterns and results in the decrease of the correct rate.

For function approach, the problem-dependent prior knowledge is the information about the goal function made the data set. When a neuron transfer function that matches better the goal function is applied, the design risk will become smaller, as our examples will show. Therefore, the design risk is enough for maximizing the problem-dependent prior knowledge for function approach.

## 7. Using a joint learning machine

There is a way to further decrease the risk, i.e., combining a huge number of GVMs designed at the same control parameter set to construct a joint learning machine, a J-GVM. The I-O sensitivity of a J-GVM is the algebraic average of these GVMs, and thus is smaller than that of a single GVM. With the smaller risk, one may expect that the J-GVM has better performance than the single ones. This is particularly useful when only a small training set is available.

GVMs composing the J-GVM are statistically identical, since they are designed by use of the same training set at the same control parameter set, while the MC algorithm makes them difference from each other. For an individual GVM, its output involves not only the information related to training samples but also random noise. The output of a J-GVM is an ensemble average of vast GVMs, the noise part is thus being offset. Due to this reason, a J-GVM constructed by a sufficient amount of GVMs can offset the training-sample independent fluctuations. Based on this property, the J-GVM needs not necessarily to be constructed using GVMs at the best control-parameter set of single GVM. For pattern recognition, application examples will show that the J-GVM constructed at a control parameter set with GVMs having a relatively big degree of I-O sensitivity may have better performance. In such a case, weight vectors have bigger freedom to extract the feature of input vectors, and therefore much more features of samples may be extracted.

## 8. Application for function approach and smoothing

We apply a $M-N-1$ GVM for function approach and smoothing. The goal functions we employed to obtain training samples are the sin function $g(x) = \sin(x)$, the sinc function $g(x) = \sin(x)/x$, and the Hermit 5th polynomial $g(x) = (63x^5 - 70x^3 + 15x)/8$. The MC algorithm is applied to train a GVM in response to samples. The cost function is the empirical risk defined by Eq. (15). The training is switched off when the empirical risk is smaller than a threshold or a given maximum number of MC steps. The latter stop condition is for smoothing noisy data sets, in which case the risk should not be decreased below the magnitude of noise.

### 8.1 Finding the best control-parameter set

We prepare three training sets from the three goal functions for function approach. Each set has 20 uniformly distributed samples $[x_i, g(x_i)]$ from interval of $x_i \in [-c_x, c_x]$ , with $c_x = \pi, 10$ , and 1 correspondingly. The stop condition is $F_e < 10^{-4}$. We prepare one noise set by adding white noise to the sinc function as $g(x) = \dfrac{\sin(x)}{x} + \zeta_i, E\zeta_i = 0, E\zeta_i^2 = \sigma^2$ for data smoothing. 100 samples with $\sigma = 0.1$ as in the reference [4] are made. The training is stopped after $10^5$N MC steps. The Gauss transfer function is adopted for all these training sets. The control parameter $c_w$ is fixed at $c_w = 10/c_x$ and as a result $c_b = 10$. The control parameter $c_\beta$ leaves to be adjustable. At each point of $c_\beta$, 500 GVMs are designed with random initializations in the available ranges of parameters. They are applied to calculate the average I-O sensitivity $\langle R_s \rangle$, the average fitting error $\langle \Theta(\Pi) \rangle$, and the design risk $E[\Pi]$.

Since the input and output layers have one neuron each, the second derivative of a GVM can be exactly calculated by applying Eq. (7). The first row of Fig. 1 shows $\langle R_s \rangle$-$R_s^g$, and the second row shows the $\langle \Theta(\Pi) \rangle$ and $E[\Pi]$, as functions of $c_\beta$ for the four sample sets. The average second derivative $\langle R_s \rangle$ is calculated over $R_s$ of 500 GVMs. The second derivative $R_s^g \equiv \| d^2 g(x) / dx^2 \|$ of the goal function is a constant and is applied as a reference line for $\langle R_s \rangle$. We see that $\langle R_s \rangle$ decreases rapidly with the decrease of $c_\beta$, which indicates that decreasing $c_\beta$ do can decrease the I-O sensitivity of the learning machine. With the further decrease of $c_\beta$, $\langle R_s \rangle$ becomes increase after a minimum. The minimum, however, is not always consistent with that of $\langle \Theta(\Pi) \rangle$. For the first and last training sets, the differences are slight, while for the second and third sets, the differences are remarkable. Particularly, for the third set, the minimum of $<R_s>$ is around $c_\beta = 1.0$ at which $\langle \Theta(\Pi) \rangle = 1.4 \times 10^{-2}$, while the minimum of $\langle \Theta(\Pi) \rangle$ appears at $c_\beta = 0.48$ with $\langle \Theta(\Pi) \rangle = 2.9 \times 10^{-3}$. Therefore, minimizing the I-O sensitivity of a learning machine does not necessarily converge to the best control parameter set.

The divergence is induced by the over minimization, i.e., the second derivative of the learning machine is minimized to a value that is smaller than that of the goal function. The overtraining is characterized by the negative interval in $\langle R_s \rangle - R_S^g$. In the cases of Fig. 1(b) and 1(c), the negative amplitudes are remarkable, and the divergence between the minima of $\langle R_s \rangle$ and $\langle \Theta(\Pi) \rangle$ are significant. In the cases of Fig. 1(a) and 1(d), $\langle R_s \rangle - R_S^g$ keeps non-negative, indicating that the overtraining has not happen yet, and the minimum of $\langle R_s \rangle$ are close to that of $\langle \Theta(\Pi) \rangle$.

On the contrary, the minimum of $E[\Pi]$ is consistent approximately to that of $\langle \Theta(\Pi) \rangle$ for each data set. For the first two sets, $E[\Pi]$ and $\langle \Theta(\Pi) \rangle$ overlap almost with each other, indicating the satisfying of $\langle \Pi(x) \rangle = g(x)$. For the last two sets, $E[\Pi]$ and $\langle \Theta(\Pi) \rangle$ show the qualitative similar dependence on $c_\beta$, indicating that the minimum of $\langle \Pi(x) \rangle$ gives still the best approach to $g(x)$. The minimum of the design risk do define the best control parameter set for function approach and smoothing.

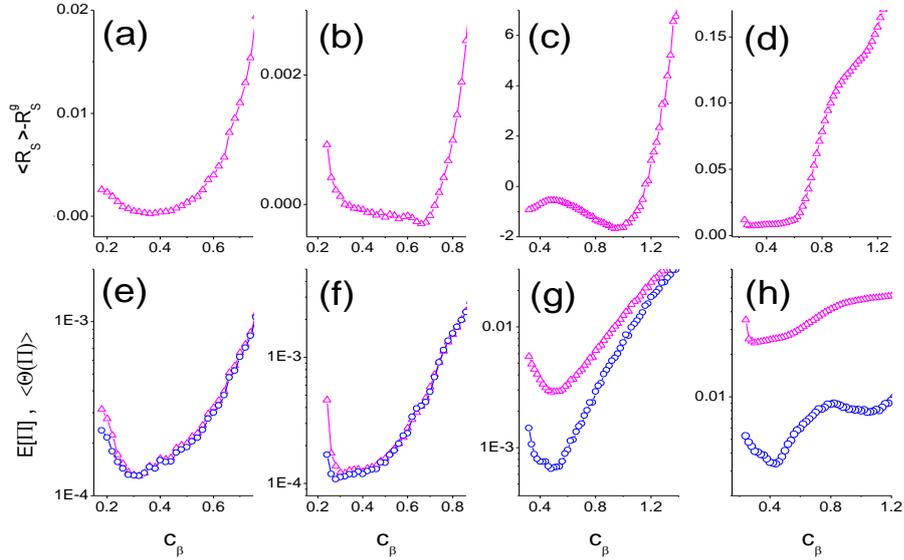

Figure 1. The first row shows $\langle R_s \rangle - R_S^g$, and the second row shows $\langle \Theta(\Pi) \rangle$ (up-triangles) and $E[\Pi]$ (circles) as functions of $c_\beta$. (a) and (e): sin function; (b) and (f): sinc function; (c) and (g): hermit polynomial; (d) and (h): sinc function with noise amplitude $\sigma = 0.1$.

We explain by example why the best control parameter set gives still the optimum fitting even when $E[\Pi]$ and $\langle\Theta(\Pi)\rangle$ are inconsistent. At the best control-parameter set, Fig. 2(a) shows two fitting curves of different GVMs for a set of noisy data of the sinc function with $\sigma = 0.2$. The two curves are almost identical, but have obvious deviation from the goal function. If more fitting curves were plotted, they also converge approximately to the same curve. Therefore, fitting curves converge to a spurious goal function instead of the real one. Since the noise biases the data set, the spurious goal function is indeed the intrinsic target function determined by the specific data set. Converging to this function instead of the real one is a reasonable result. With the decrease of the noise amplitude, the induced bias decreases, and the spurious one will be close to the real one, as Fig. 2(b) shows. This situation may occur for noise-free training samples as well. A finite sample set cannot exclusively determine a goal function. The neural transfer function applied, together with the architecture of GVM, determine a spurious goal function. This is the case of Fig. 1(g).

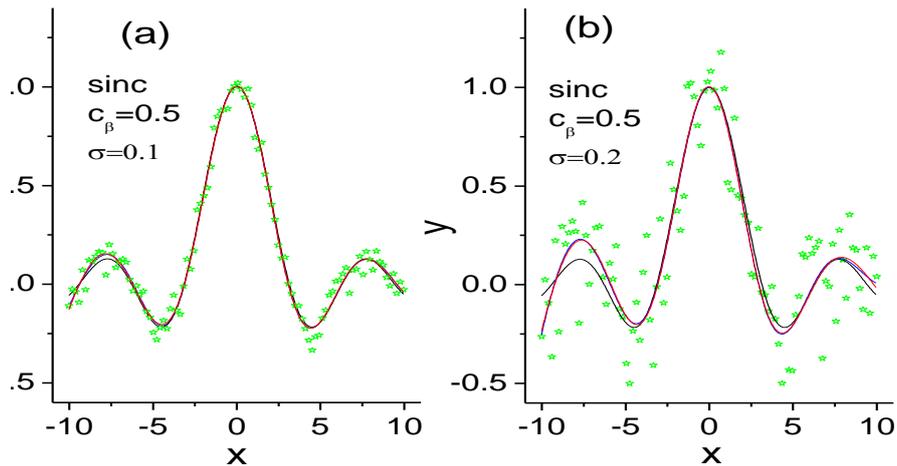

Figure 2. Function smoothing for samples with noise intensity of $\sigma = 0.1$ (a) and $\sigma = 0.2$ (b). 100 samples (stars) are distributed uniformly in the interval of [-10,10]. The goal function is the 1D sinc function (black line). In the plot, two fitting curves are shown, which are almost overlapped with each other.

## 8.2 Improving the fitting by increasing the learning machine size

We see from Table I, increasing samples can greatly increase the fitting precision. This is generally true for also conventional methods. Here we emphasize that we can also improve the fitting precision by increasing hidden-layer neurons. Increasing the hidden-layer neurons from 100 to 1000, the fitting precision may be further improved by up to threefold. This is a remarkable difference from conventional methods. The BP method follows the 'Occam razor principle' to pursue neural networks with as smaller as possible size. If a hidden layer with 10-100 times of samples is applied, serious over-fitting must be arisen. For the SVM method, the hidden-layer neurons are limited by the amount of samples, which could not exceed the number of samples.

Table I : Fitting precision vs. sample amount and machine size

| Samples | 10 | | | 20 | | |
|---|---|---|---|---|---|---|
| Data | Sin | sinc | hermit | sin | sinc | hermit |
| GVM(1-100-1) | $4.5 \times 10^{-4}$ | $4.5 \times 10^{-3}$ | $2.5 \times 10^{-2}$ | $1.3 \times 10^{-4}$ | $1.2 \times 10^{-4}$ | $2.9 \times 10^{-3}$ |
| GVM(1-1000-1) | $1.3 \times 10^{-4}$ | $2.6 \times 10^{-3}$ | $1.0 \times 10^{-2}$ | $1.2 \times 10^{-4}$ | $1.1 \times 10^{-4}$ | $1.0 \times 10^{-3}$ |
| J-GVM(1-100-1) | $5.3 \times 10^{-5}$ | $2.4 \times 10^{-3}$ | $2.5 \times 10^{-2}$ | $1.0 \times 10^{-5}$ | $3.6 \times 10^{-5}$ | $2.9 \times 10^{-3}$ |
| Spline | $3.2 \times 10^{-4}$ | $1.5 \times 10^{-2}$ | $8.2 \times 10^{-2}$ | $1.5 \times 10^{-5}$ | $1.1 \times 10^{-3}$ | $1.3 \times 10^{-2}$ |

## 8.3 Fitting by a J-GVM

To obtain GVMs, the training is stopped when $F_e \le 10^{-4}$ for function approach examples. Therefore, the fitting precision of a GVM could not be beyond this threshold. Table I shows that applying a J-GVM can usually obtain a better result than a GVM. The J-GVM is constructed by the 500 GVMs trained at the best control parameter $c_\beta$ where $E[\Pi]$ takes the minimum. For certain data sets the precision can be improved by even one order, which is significantly higher than the empirical risk.

## 8.4 Comparison to conventional algorithms of function approach

Table I also presents results using the widely applied spline algorithm for function approach. For the last two sets, GVMs get obviously better performance. For the data of sin function, the spline algorithm reaches the same precision. However, this is because we apply the training stop condition of $F_e \le 10^{-4}$, which limits the precision of the fitting. If one changes the stop condition to $F_e \le 10^{-5}$, one can further improve the precision of GVMs to the order of $10^{-5}$. The J-GVM is constantly better than the traditional algorithm for each data set.

As to comparing to the SVM, we would like to mention the example shown in the text book of Vapnik [4]. In that example, 100 samples for the sinc function are applied as the training set. When choosing 14 samples from this set to be support vectors, the fitting curve already shows big diverges that can be obviously seen. In our case, even for the training set with only 10 samples, one can approach a precision of $<\Theta(\Pi)> = 4.5 \times 10^{-3}$ using a GVM. The fitting curve is already visually indistinguishable from the goal function.

## 8.5 Training time

For the sake of comparison, the training time is calculated by the CPU time of commonly used personal computer (specifically with 2.0 GHz). Figure 3 shows the average training time of a GVM as a function of $c_\beta$ in the case of the sinc goal function with 20 samples. The training is achieved if the cost function is decreased below $F_e = 10^{-4}$. It can be seen that the training time increase rapidly with the decrease of $c_\beta$. The training time may increase exponentially when the GVM becomes too small. We have checked that for $N \leq 20$, the training fails due to the condition of $F_e < 10^{-4}$ cannot be approached within a reasonable training time. On the contrary, it decreases with the increase of the machine size. Together with the fact that large machines may improve the fit precision, our strategy thus prefers larger machines than small ones. Over-fitting problem of large machines can be suppressed by controlling the design risk.

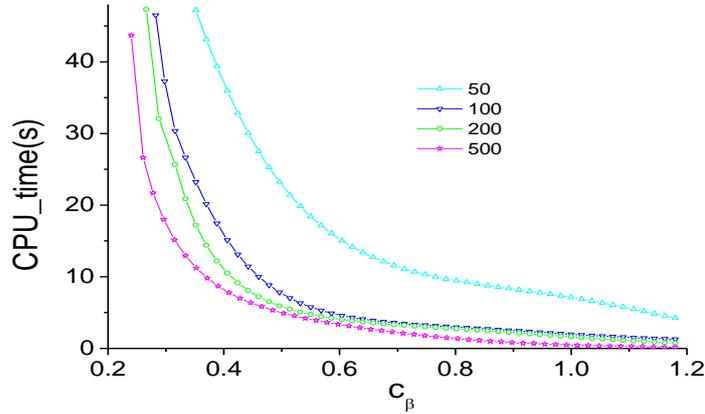

Figure 3. Training time as functions of the machine size.

## 8.6 Improving the fitting by proper neural transfer functions

The BP method is derived with the sigmoid-like neuron transfer function and thus has little freedom for choice of the transfer functions.

Choosing the transfer function (kernel function) remains a vexing issue in SVM method. One usually needs to search for different function for different problem. As for fitting the sinc function, a complex form of kernel function $f_i(x, x^i) = 1 + x \wedge x^i + (1/2)|x - x^i|(x \wedge x^i)^2 + (1/3)(x \wedge x^i)^2$ is employed, where $x^i$ represents a support vector [4] and $x \wedge x^i$ represents the inner product between the input vector and the support vector. In our examples, favorable fittings are obtained with the Gauss transfer function for either data set.

However, when the transfer function matches the feature of the goal function better, one may obtain a better result. Figure 4 shows the fitting results using the Gauss, sigmoid, and polynomial neuron transfer functions for data sets of the sin, sinc and Hermit 5th polynomial. In each set there are 10 sample points. The polynomial transfer function is defined by a six order polynomial $f(z) = z^6$.

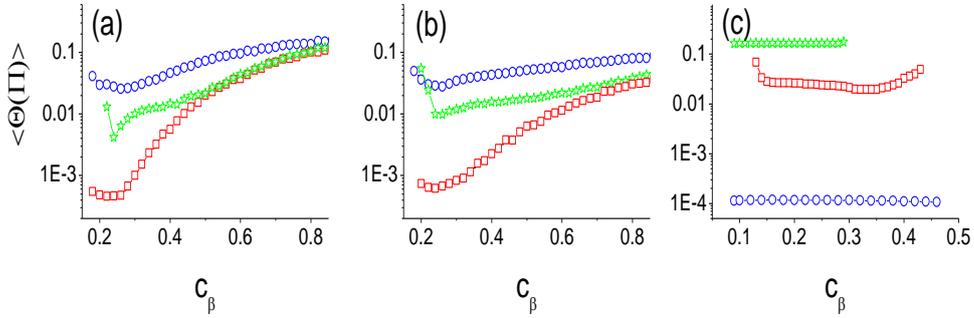

Figure 4. The fitting precision as a function of $c_\beta$ by use of the Gauss (a), Sigmoid (b), and polynomial (c) neural transfer functions. The squares, stars, and circles are for data sets from the sin, sinc, and Hermit polynomial goal functions, correspondingly.

We see that fitting with the Gauss transfer function can approach better results for all the three data sets than with the sigmoid transfer function, but the difference is not remarkable. Applying the polynomial transfer function can make a big difference. For the data set of the sinc function, the best precision is about $\langle\Theta(\Pi)\rangle = 0.16$. Indeed, even the empirical risk can only be minimized below to $F_e = 0.10$ for this set. For the data set of the sin function, a precision of $\langle\Theta(\Pi)\rangle = 0.02$ is approached, but is still quite worse than those using another two transfer functions. For the Hermit polynomial function, however, a much high precision is achieved. With the training stop condition of $F_e \leq 10^{-4}$, the fitting precision remains below $\langle\Theta(\Pi)\rangle = 1.1 \times 10^{-4}$ as Fig. 4(c) indicated. With $F_e \leq 10^{-6}$, one can achieve $\langle\Theta(\Pi)\rangle = 1.1 \times 10^{-6}$. These facts imply that the goal function is recovered with remarkable precision. We have checked that the high-precision fitting can always achieve for this data set if applying a transfer function of $f(h) = h^n$ with $n \geq 5$. The perfect precision obviously comes from the fact that the polynomial transfer function matches the feature of this data set well. Similar discussion is applicable to explaining the results in Fig. 4(a) and 4(b), where the Gauss transfer function marchers the data sets better than the sigmoid one. The more favorable neural transfer function can be chosen following the design risk criterion or the prior knowledge of the data source.

## 8.7 Applying as a universal fitting machine

For practical applications, one needs not necessarily to search for the best-control parameter set. One may have noticed that the best fitting achieves around $c_\beta = 0.5$ for all of the data sets. With this parameter value and a 1-200-1 GVM with Gauss transfer function, we find the fitting can be done well for several more complex goal functions also, as Fig. 5 shows. The first data set comes from the sinc function in the interval of [-20,20], and the second from the sin function in the interval of $[-5\pi, 5\pi]$. The third set is from the Hermit 7th polynomial, $g(x) = (429x^7 - 693x^5 + 315x^3 - 35x)/16$. The last set is from a square wave in [-10,10]. For each set, only 20 samples are used.

This is due to that we properly rescale the input of data sets. We set $c_w = 10/c_x$ and thus $c_b$ is fixed at $c_b = 10$, which results the local field $\bar{h}_i$ distributing in the same interval. In this case, the best $c_\beta$ should be roughly identical. We therefore choose a proper value, $c_\beta = 0.5$, according to Fig. 1 to fulfill the purpose of fitting various data set. Sometimes, the data set may show a more complex flexibility, and the training may be not achieved (the empirical risk cannot be decreased below the target threshold of $F_e$). In this case, one can still use $c_\beta = 0.5$ but allow the computer to increase the hidden-layer neurons until the training is achieved, since large learning machines have large flexibility. With such a strategy, for usual application of function approach, the calculation of $E[\Pi]$ using a large number of GVMs can be avoided. This kind of calculation is required only when the higher fitting precision is essential.

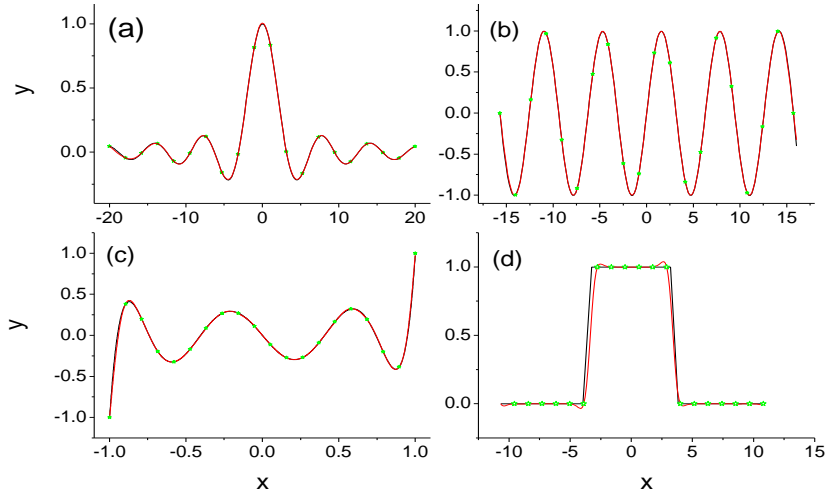

Figure 5. Fitting different data set using a *1-200-1* GVM with the Gaussian transfer function. (a) The sinc function in [-20,20]. (b) The sin function in $[-5\pi, 5\pi]$. (c) The hermit 7th polynomial in [-1,1]. (d) A piecewise step function in [-10,10]. There are 20 samples in either sets.

The algorithm can be implemented directly to high dimensional fitting. The SVM method gives a desirable fitting precision by choosing 153 support vectors from $20 \times 20$ samples [4] of the two-Dimensional sinc function $g(x, y) = \sin(\sqrt{x^2 + y^2}) / \sqrt{x^2 + y^2}$. Applying our strategy to design a 2-1000-1 GVM, Fig. 6 shows that a better result is achieved by use of $10 \times 10$ sample points only.

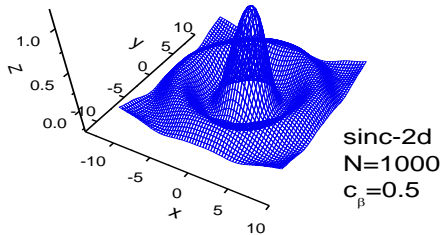

Figure 6. Fitting 2D sinc function using 10x10 samples with a *2-1000-1* GVM.

## 8.8 The role of bias

Here we provide examples to explain the role of the neuron bias. The I-O sensitivity of a fitting curve should be the same as to that of the goal function and be independent on the specific value of the input coordinate. However, if set $c_b = 0$ it has $f_i^{''}(0) = 0$ at the origin for any data set and for all those neural transfer functions, as pointed out in section 4. The I-O sensitivity at the origin is thus also equal to zero. The control fails at this point. Figure 7 shows the results that using GVMs with $b_i = 0$ to fit the data sets of the sinc and sin function. Around the origin, one can clearly see big deviations, indicating that at the origin the fitting curve do is stiff and thus the fitting is difficult to be fulfilled.

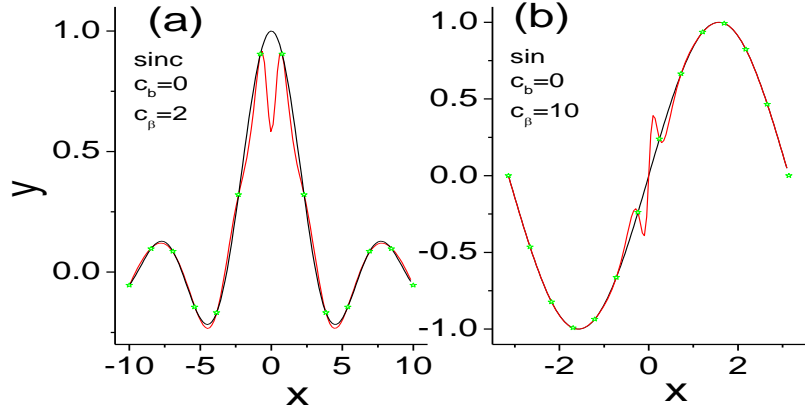

Figure 7. The fitting with $b_i = 0$ for the sinc function (a) and the sin function (b).

To disentangle this problem, we need to decouple the dependence of $f_i^{''}$ on the value of inputs, and so as to guarantee the sensitivity of a fitting curve is determined by the intrinsic behavior of the data set instead of values of inputs. For this purpose, we set $c_b = max \sum \overline{w}_i x_i$, in which case the random initializations of $\overline{w}_i$ and $b_i$ led to the distribution of $f_i^{''}$ insensitively depend on the particular value of inputs.

# 9 Pattern Recognition

We perform a standard handwritten digit recognition task to demonstrate how to design the GVM for pattern recognition. The dataset MNIST [23] has 60000 training samples and 10000 test samples. Each sample is represented by a $28 \times 28$ dimensional vector. To perform this task, the BP method trains a $28 \times 28 - N - 10$ multi-classifier[24,25], while the SVM method usually designs ten binary-classifiers [26].

The GVM has the $28 \times 28 - N - 10$ architecture. The original data $x_i^{\mu}$ takes integer of $(0,255)$. We rescale the component by $0.1 * x_i^{\mu}$ as inputs. The output target vector $y^{\mu}$ encodes the digit $v$, $v \in (0,...,9)$, which is defined by $y_l^\mu > 0$ for $l = \nu + 1$ and $y_l^\mu \leq 0$ otherwise. For sake of simplicity, we set $c_w = 1$ and $c_b = 100$, and set $w_{il}$ with random binary numbers. These settings keep unchanged in the subsequent design process. Till section 9.7, we only apply the first 1% MNIST training samples to be the training set.

## 9.1 Finding the best control-parameter set

To decrease the I-O sensitivity, we can decrease $c_\beta$ at a fixed *d*, or increase *d* at a fixed $c_\beta$. In doing so, we can search the best control-parameter set along only one parameter axis. At a control-parameter set, 500 GVMs are applied to calculate the average recognition accuracy $\langle \Pi \rangle$, the average second derivative $\langle R_s \rangle$, and the design risk $E[\Pi]$. Hide-layer neurons are modeled by the Gauss neural transfer function, and the applied cost function is $F_2$ with $F_2 < 1$. To estimate the second derivative, only $\gamma_{jj}^i$ of the hidden-layer neurons are involved in Eq. (8), otherwise, the calculation should be a hard task. We have checked by full calculation of $\gamma_{jk}^i$ on a small training set and found no qualitative difference.

We first show the dependence of $\langle \Pi \rangle$, $\langle R_s \rangle$ and $E[\Pi]$ on $c_\beta$ with *d* fixed. The circles and triangles in Fig. 8 show the results with *N=1000* and *d=30*, and *N=3000* and *d=100*, respectively. We see that $\langle R_s \rangle$ and $E[\Pi]$ decrease monotonously with the decrease of $c_\beta$. The recognition rate $\langle \Pi \rangle$ increases rapidly with the decrease of $c_\beta$ at first. After the turning point of $c_\beta \approx 0.005$, it turns to decrease.

Therefore, the best control-parameter set is not at the minimum of neither $\langle R_s \rangle$ and $E[\Pi]$. This is just an example that extremely pursuing the common knowledge may induce the divergence from the problem-dependent knowledge. In this case, the best control-parameter set is determined as a balance between a high recognition rate and an acceptable design risk. In this example, the turning point of the average rate can define the best control parameter set since at which the design risk is acceptably low.

Figure 8 reveals more. Firstly, applying large machines cannot only increase the recognition rate but also decrease the design risk. At the turning point, the average recognition accuracy is 89.3% for N=1000 and 90.1% for N=3000, while the design risk is about 0.2% and 0.1%, correspondingly. Secondly, it reveals that the turning point of $c_\beta$ is insensitive to the GVM size. Either for N=1000 and N=3000, it appears around $c_\beta = 0.005$. We can thus fix the parameter $c_\beta$ to this value in our following studies.

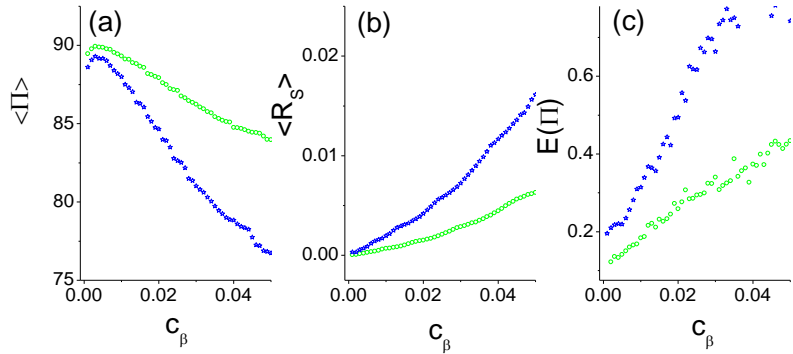

Figure 8. The average recognition rate (a), the structural risk (b), and the design risk (c) as functions of control parameter $c_\beta$. The stars for $d = 30$ and $N = 1000$, and the circles for $d = 100$ and $N = 3000$, respectively.

We then study the dependences on $d$ at $c_\beta = 0.005$. Fig. 9 shows the results for *N=3000*. We see that initially $\langle \Pi \rangle$ increases rapidly with the increase of *d*, but becomes decrease after a turning point around

*d=120.* The design risk decreases monotonously with the increase of *d*. Thus, over maximizing the separating margin may result in the overtraining for this data set. The second derivative $\langle R_S \rangle$ indeed increases slowly. This is because, though the parameter $c_\beta$ is fixed, the MC adaption may induce $\beta_i$ concentrating slightly towards the boundaries of the specified interval and thus increases $\langle R_S \rangle$.

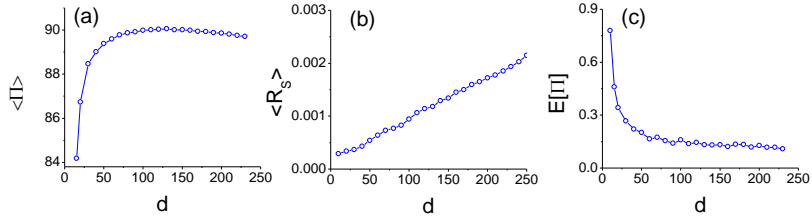

Figure 9. The average recognition rate (a) the structural risk (b) and the design risk (c) as functions of control parameter *d* at $c_\beta = 0.005$ and $N = 3000$.

In principle, the larger the separating margin, the big the probability that a variation of a sample being classified into the same class. This is, however, true when the test set can be considered to be random variations of training samples. Below we illustrate this guess by examples. We construct two test sets using virtual samples. The first one is created by adding random noise to input vectors of the first 1% samples as $x_i = x_i^\mu + \zeta_i$, $E\zeta_i = 0, E\zeta_i^2 = \sigma^2$ with $\sigma = 80$. For each sample, 10 virtual samples are made and thus totally 6000 samples are involved in this set. The second one is obtained by shifting each of the 1% sample patterns with 2 pixel units to adjacent positions, which gives totally 4800 samples then. Figure 10 show the average recognition accuracy measured on these tow test sets for GVMs with N=3000 as a function of the control parameter *d*. For comparison purpose, the result on the original test set is also shown. Other control parameters keep same as in Fig. 9. We see that with the increasing of *d*, $\langle \Pi \rangle$ of the first set increases monotonously, while for another two sets the over-training effect appears after the same turning point.

These results confirm that the maximum-margin strategy applied by the SVM method is correct quantitatively when the real patterns can be considered as random variations of training samples. In other wards, it is correct generally for maximizing the common prior knowledge. For practical applications, as for handwritten digits here, patterns cannot be considered to be completely random variations since they are restricted by the particular geometry.

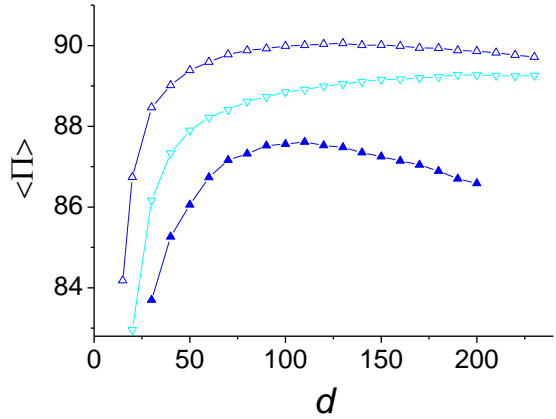

Figure 10. The average recognition rate as a function of $d$ at $c_\beta = 0.005$ and $N = 3000$ for the original test set (up-triangles), the test set of random variations (down-triangles), and the test set of shifted samples (solid triangles).

The dependence of $\langle \Pi \rangle$ on $d$ for the test set of shifted virtual samples is similar to that of using the original test set. This fact can be interpreted as that the shift operation keeps the geometry of digit patterns. Because it gives the same turning point as using the original test set, one can apply this set to find the best control parameter set and apply the original test set also to train the learning machine, in which way the samples may be maximally utilized.

## 9.2 Improving the recognition rate by increasing the machine size

Increasing machine size can extract more features of samples, and thus can increase the generalization ability. In more detail, each weight vector extracts information of samples from a different angle, and thus the more the weight vectors, the more features of samples can be extracted. Figure 11 shows the dependence of the average recognition accuracy on the machine size, which indicates that increasing the size can improve the recognition rate monotonously, though the benefit may become saturated when the size is adequate.

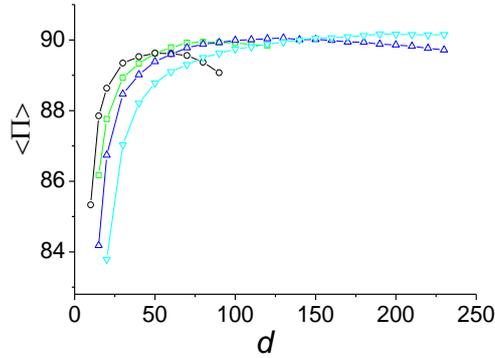

Figure 11. The dependence of the average recognition rate on the machine size. The hollow circles, squares, up-triangles and down-triangles are for GVMs with $N=1000$、 $2000$、 $3000$、 $6000$, correspondingly.

The figure shows that the best control parameter set depends also on the neurons number $N$. In section 9.1 we have shown that the best value of $c_\beta$ is insensitive to other parameters, we can therefore search the space $d-N$ for the best control parameter set by fixing $c_\beta$ at $c_\beta = 0.005$.

## 9.3 The role of neuron transfer function and cost function

In Figure 12, We show results with four pairs of combinations of cost function and neuron transfer function, $F_2$-Gauss (circles), $F_1$-Gauss (stars), $F_2$-sigmoid (up-triangles), and $F_1$-sigmoid (down-triangles), respectively. The GVM size is *N=3000*. The training stop condition is $F_1 < 10^{-4}$ and $F_2 < 1$ for the two cost functions respectively. Obviously, the $F_2$-Gauss combination gives the best result, indicating that Gauss type functions match better the nature of the spatial pattern. The result of the ReLU neuron transfer function with cost the Gauss cost function $F_2$ is also shown as a reference. We see that it is not the best one for this recognition task.

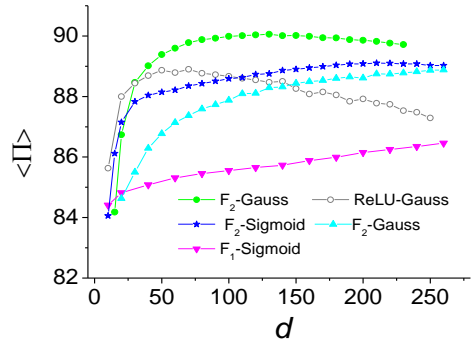

Figure 12. The dependence of the average recognition rate on transfer functions and cost functions.

## 9.4 Using a J-GVM

At a given point of *d* we design 500 GVMs of size *N*=3000 with the Gauss transfer function. The J-GVM is constructed by these GVMs. We use $\Pi'$ to represent the recognition rate of the J-GVM. Figure 13 shows $\Pi$, $\langle \Pi \rangle$ and $\Pi'$ as functions of *d*. Figure 13(a) is for applying the Gauss transfer function and Fig. 13(b) applying the polynomial transfer function with n=7. The cost function is $F_2$ for both cases.

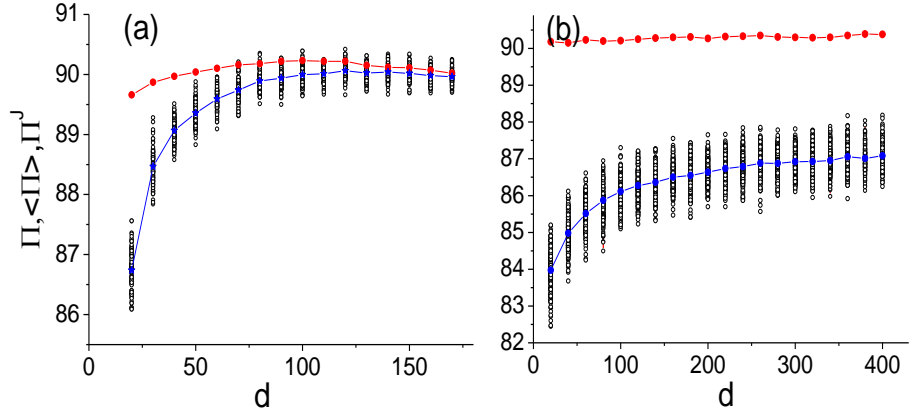

Figure 13. $\Pi$ ( circles), $\langle\Pi\rangle$ ( solid stars) and $\Pi^J$ (solid circles) as functions of *d*, (a) for the Gaussian transfer function and (b) for the polynomial transfer function respectively.

We see that, besides having a high value, the recognition rate is relatively insensitive to the control parameter. In Fig. 13(a), J-GVMs designed in $d \in (50,150)$ have closed recognition rate, and in Fig. 13(b), the recognition rate seems just slightly dependent on *d*. This property is an essential advantage of a J-GVM based on which a complete searching for the best control parameter set is avoided. The reason why a J-GVM is insensitive to the control parameter is as follows. The J-GVM has a small design risk even at the control-parameter set that individual GVM has big I-O sensitivity, since outputs of GVMs can be considered as random fluctuations around the target output and thus are offset with each other. Therefore, even the I-O sensitivity of an individual GVM is relatively big, the I-O sensitivity of the J-GVM may still be sufficiently small if only the GVMs are enough.

Figure 13(a) reveals another phenomenon, i.e., the best control parameter *d* of a J-GVM is smaller than that of individual GVMs. As will also see in the next section, this is a common property of a J-GVM. The reason is that the recognition rate is determined by features being extracted from samples, and GVMs with bigger risks have bigger freedom in extracting features. If only the risk is suppressed, the J-GVM constructed by GVMs having big I-O sensitivity will show this superiority.

The polynomial transfer function shows a noteworthy feature. Though the recognition rates of GVMs are quite low, the rate of the J-GVM is even higher than that of using the Gauss transfer function. The reason may be due to that this transfer function matches the nature of digit patterns better, i.e., because the change of the gray degree of digit patterns is steep, higher-order polynomial transfer functions fit this feature well. Nevertheless, since the second derivative $f_i^{''}(z) = n(n-1)z^{n-2}$ is very big, individual GVMs with this transfer function have bad performance. When applying a J-GVM, the risk is suppressed by the ensemble average, and the advantage that high-order polynomial transfer functions emerge.

## 9.5 Improving the performance by proper pretreatment of samples

As explained in section 9.1, the maximum-margin strategy is generally applicable when the test samples can be considered as random variations of training samples. For handwritten digits as well as the usual spatial patterns, variations could not be considered as random. The particular geometry of spatial patterns excludes most of the random variations. To create a virtual sample set following the geometric nature of patterns is a means to avoid the excessive generalization. We have constructed a spurious sample set in section 9.1 by shifting each of the first 1% MNIST sample patterns with 2 pixel units to adjacent positions. It is applied as a test set there. Here we apply it to be a training set. Figure 14 shows that the recognition rate on the test set is significantly improved comparing to that using the original 1% samples.

Certain simple retreatments of input vectors may be also effective. For example, we smooth the first 1% samples by using a Gauss convolution with unit standard deviation and applied them to train the learning machine, the recognition rate is further improved, see Fig. 14. The more effective way of encoding spatial information is the Gradient-based feature extracting technique. A 200-dimensional numeric feature vector encoding eight direction-specific $5 \times 5$ gradient images is calculated for each sample using this technique, as described in [22]. This is one of the three top-performing representations in the reference and is called as *e-grg*. Applying the first 1% samples pretreated by this technique we approach a much high recognition rate shown also in Fig. 14.

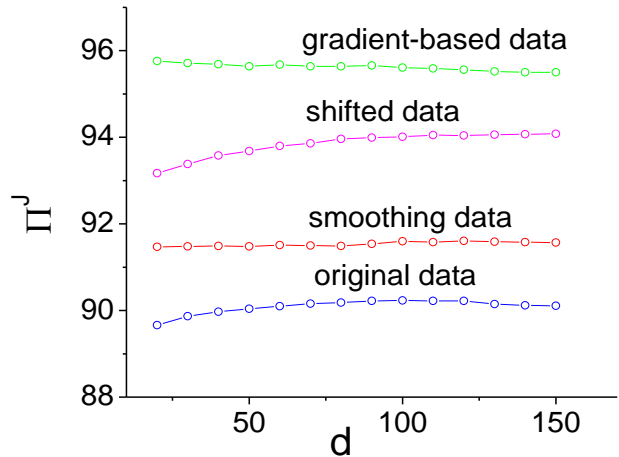

Figure 14. The recognition rate as a function $d$ at $c_\beta = 0.005$ and $N = 3000$ for the J-GVM designed by different training set constructed by the original 1% MNIST data set. A J-GVM is constructed by $100$ GVMs.

## 9.6 The highest record on the data set

Similar to other training methods, increasing training samples can increase the recognition rate. Figure 15 show the results of using the first 10% and all of the MNIST samples respectively. In the first case 50 GVMs, and in the last case 10 ones, are trained respectively at each parameter point, and the J-GVM is constructed with these GVMs. In both cases, the Gauss transfer function is applied and the GVM size is fixed at N = 6000. The cost function $F_2$ with training termination condition $F_2 < 1$ is applied. **In the training, the normalized gray-scale images are directly used so as to purely compare algorithms themselves, with getting rid of the influence resulted by pretreatment techniques**. It can be seen that using all of the training samples the record is beyond those using the BP method with error rate 1.5% [25], the SVM with error rate 1.4% (By combining 10 one-vs-rest binary SVMs and building a ten-class digit classifier) [26] and milt-layer conventional neural network with error rate 1.25% [25]. The last record is obtained with a complex five-layer hierarchical model. Therefore, in the case of applying the original training set, our record is competitive.

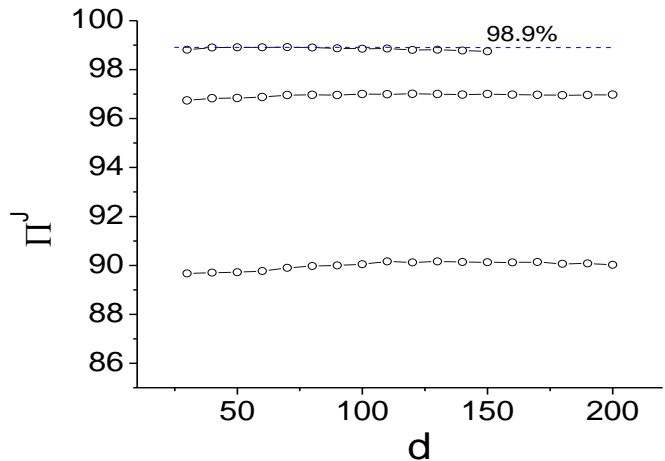

Figure 15. The recognition rate as a function $d$ at $c_\beta = 0.005$ and $N = 6000$ for the J-GVM designed by the original first 1%, 10%, and the complete set of MNIST data set, correspondingly.

# 10 Classification

Classification is a special example of pattern recognition. The Wisconsin breast cancer database was established [27] in 1992 with 699 samples. As usual, first 2/3 samples are applied as the training set, and the remains as the test set. The inputs are 9 dimensional vectors, with components represent features from microscopic examination results, and are normalized to take value from [0, 10]. The output indicates the benign and malignant patients.

This task can be done by a GVM with two neurons in the output layer. A patient is classified into benign if the output of the first neuron is bigger than that of the second one, otherwise malignant. We first study $\langle \Pi \rangle$ and $E[\Pi]$ as a function of the control parameter $d$ with other parameters keeping fixing at $c_\beta = 1$, $c_w = 1$, $c_b = 10$ and $N=200$. The weights $w_{il}$ in the output lay are fixed at a set of randomly initialized binary numbers. At each point of $d$, 500 GVMs are used to calculate $\langle \Pi \rangle$ and $E[\Pi]$.

Figure 16 presents the results. The representations of the symbols are: up-triangle for $F_1$- sigmoid combination; down-triangle for $F_1$-Gauss combination; star for $F_2$-sigmoid combination; circles for $F_2$-Gauss combination. Stop conditions are $F_1 < 10^{-3}$ or $F_2 < 1$ correspondingly. When the stop conditions cannot be met within a preset maximum training time, the search alone $d$ is ceased.

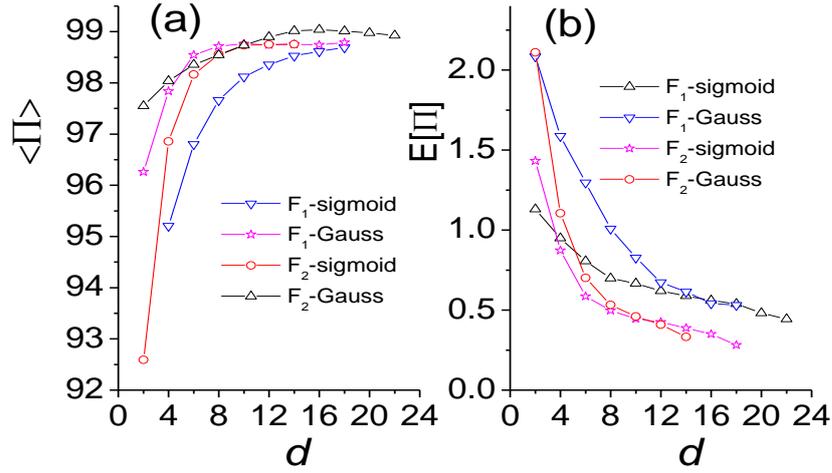

Figure 16 $\langle\Pi\rangle$ and $E[\Pi]$ as a function of $d$.

We see that the best result is given by the cost function $F_1$ with the sigmoid transfer function. This fact indicates that the monotonic functions match the nature of this sample set well. The reason is that for a component of such a sample vector, small value means the normal, while larger one represents the abnormal.

With the $F_1$- sigmoid combination, the maximum of $\langle\Pi\rangle$ is observed around $d \sim 16$, after which it becomes decrease slightly. This phenomenon might be explained as that the data from the medical examination could be regarded approximately as random variations of standard samples. The microscopic examination may induce random errors of measurement, and meanwhile biochemical indexes themselves may be influenced in a complex way by prompt accidental events of a patient.

Figure 16(b) shows that $E[\Pi]$ decreases monotonously with the increase of $d$. The essential feature explored here is the big uncertainty. For example, in using the $F_1$-sigmoid combination, $E[\Pi]$ is about $\pm 0.5\%$ even at $d=16$ where the average correct rate takes the maximal value of $99.0\%$. The uncertainty is therefore a serious problem.

The J-GVM provides an effective way to remedy this drawback. Fig. 17 shows the distribution of the recognition rate of GVMs as a function of the control parameter $d$. At each point of $d$, the correct rates of 500 GVMs designed with the $F_1$-sigmoid combination are shown as stars. The average correct rate of GVMs and the correct rate of the J-GVM constructed using these GVMs are also shown in the figure as triangles and circles respectively. Fig. 17(a) is for $N=200$ and Fig. 17(b) for $N=500$. In the case of N=200, the maximal average correct rate is 99.01%. In an interval of $d \in [4,8]$, the rate of the J-GVM keeps at 100%. With more big GVMs, N=500, the maximum average correct rate approaches 99.30% and in a wide interval of $d \in [4,22]$ the correct rate of the J-GVM keeps at 100%. This fact again indicates that bigger machines are more favorable. As to the records of the correct rate, our results are superior to previous studies [28,29], where a record of 97.5% is approached by a SVM-method based learning machine.

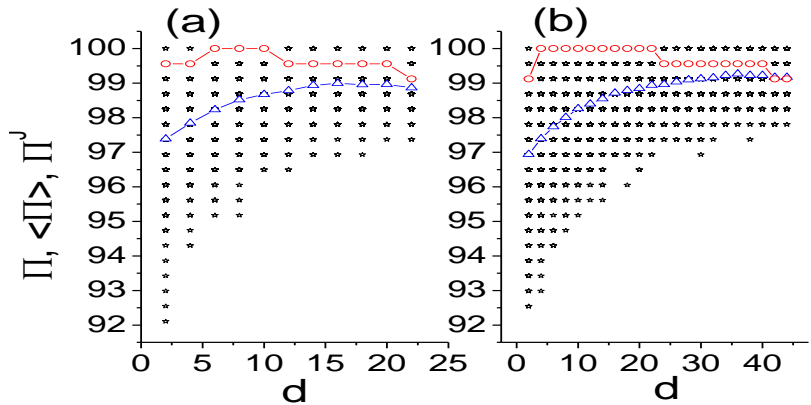

Figure 17. $\Pi$ ( circles), $\langle \Pi \rangle$ ( solid stars) and $\Pi^J$ (solid circles) as functions of $d$, (a) for N=200 and (b) for N=500.

Figure 17 also explores the drawback of applying the individual learning machine to be the performing machine. Even in the parameter region with low average correct rate, as at $d= 2$ in Fig. 17, certain GVMs may approach the correct rate of 100%. However, at the same parameter, another one may approach only about 92%. Therefore, the correct rate on the test set is not the right indicator of learning machine performance. The higher record may be just a fluctuation. When applying it to real patients, one cannot expect it still remains the high correct rate. The learning machine with a 92% record may not be necessarily worse than that with a 100% record for real application; they have the same expectation value for unlearned data. The J-GVM avoids such kind of uncertainty.

## 11 Summary and Discussion

(1) We develop a MC algorithm to gain the correct response to the training set. Applying this algorithm, one can obtain three-layer learning machines with either continuous or discrete parameters. Not only the weight vectors, but also the neuron transfer coefficients become adjustable internal parameters. Indeed, different type of neuron transfer functions can be applied simultaneously in a learning machine. This algorithm enables us to give up support vectors and replace them by general weight vectors. For small training-set problems, support vectors are limited by the number of samples. The general weight vectors have no such a limitation. Using enough weight vectors, the features of input vectors can be maximally extracted. An of important conclusion obtained from these facts is that the simple MC algorithm may be practicable for designing multi-layer learning machines, by properly designed algorithm the training efficiency is acceptable for usually applications.

(2)    We classify the prior knowledge into common and problem-dependent classes, and suggest corresponding strategies to incorporate

them into the learning machine. The common prior knowledge involves that a learning system should has small I-O sensitivity and small design uncertainty. We find that decrease the design risk can suppress the I-O sensitivity, and therefore we apply the DRM principle to incorporate the common prior knowledge. We have shown that the design risk can be applied as the unique quantitative indicator for function approach and smoothing. We further clarify that maximizing the common prior knowledge maximizes the generalization ability when real patterns can be considered as random variations of training samples. In this case, the samples are not restricted by geometric symmetry. When be constrained with particular geometric symmetry, pursuing to classify all random variations of a sample to be its class is extravagant. The specific geometric symmetry is a typical problem-dependent prior knowledge, as the interpretation of input vectors of samples. Maximally pursuing the common prior knowledge and problem-dependent prior knowledge sometimes may induce contradictory. Extremely maximizing the common knowledge by maximizing the separating margin as SVM method does can result overtraining due to the divergence to the geometric symmetry. To monitor the design in this situation, we apply also the average correct rate to be another performance indicator for pattern recognition and classification. A more proper neuron transfer function and/or cost function can maximize the problem-dependent prior knowledge, and constructing spurious samples following the natural geometric symmetry of samples and/or incorporating the geometric information into input vectors are also means for this purpose.

(3) We have shown that the second derivative of a neuron is determined by the multiple of weights, neuron transfer function coefficients, and the second derivative of the neuron transfer function. By applying neuron transfer functions having bounded second derivative, the I-O sensitivity of neurons can be controlled by control parameters specifying the available ranges of weights and neural transfer coefficients. As a linear combination of individual neurons, the I-O sensitivity of the learning machine can thus be controlled by these control parameters.

Instead of finding the best machine according to the test result on the test set as conventional design methods do, we search for the best control parameter set along the direction with decreasing I-O sensitivity. Each GVM designed at the best control parameter set has the same expectation performance for real patterns, and thus can be equally applied as the performing learning machine. We can instead to apply the J-GVM constructed by a sufficient amount of GVMs designed at a control parameter set to be the performing machine. The output of a J-GVM is the ensemble average of these GVMs. The J-GVM usually has better performance since it has more small risks.

(4) We emphasize that the superiority of our method is particularly for small sample-set problems. In this case the GVM as well as the J-GVM have obviously higher performance than conventional learning machines. Even for handwritten digit recognition with 60000 samples, we can beat conventional ones if applying only the original gray-scale images without any pretreatment so as to fairly compare the training strategy itself.

(5) Our method paves a way for wide applications. Besides it can be extended to many other traditional tasks of learning problem, such as the time-series prediction, our algorithm may induce new applications. For example, after proceeding the Monte Carlo adaptation for a proper period, the local fields $h_i^\mu s_i^\mu$ will distribute around $s_i^\mu h_i^\mu = d$ . Then those examples with $h_i^\mu s_i^\mu << d$ may represent 'bad examples'. The figure below shows 20 such instance. One can see that, the third sample and the last one for example, no one could recognize them as '3' and '4' as the tags indicated. To pick out these bad examples, the test set is not used. It may be an instance of the so-called transductive inference [4]. Picking up bad instances might have practical importance for certain problems, such as finding those misdiagnosed patients from the training set.

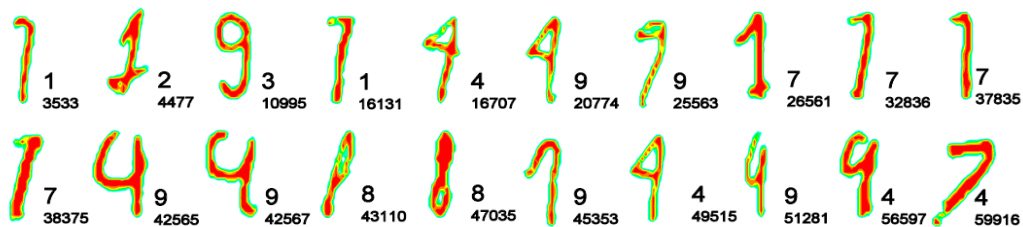

Figure 18, 'Washing out' the bed samples. The digit patterns are from the MNIST data set. The above numbers of the subscript are target digits, below ones are the sequence number of patterns in the data set, correspondingly